# Explainable Reinforcement Learning on Financial Stock Trading using SHAP

Satyam Kumar[a,b], Mendhikar Vishal[a,b], Vadlamani Ravi[a*]

[a] Center of AI and ML, Institute for Development and Research in Banking Technology (IDRBT), Castle Hills, Masab Tank, Hyderabad 500057, India

[b] School of Computer and Information Sciences (SCIS), University of Hyderabad, Hyderabad 500046, India

learnsatyam@gmail.com; mendhikar.v@gmail.com; vravi@idrbt.ac.in

**Abstract**

Explainable Artificial Intelligence (XAI) research gained prominence in recent years in response to the demand for greater transparency and trust in AI from the user communities. This is especially critical because AI is adopted in sensitive fields such as finance, medicine etc. where implications for society, ethics, and safety are immense. Following thorough systematic evaluations, work in XAI has primarily focused on Machine Learning (ML) for categorization, decision, or action. To the best of our knowledge, no work is reported that offers an Explainable Reinforcement Learning (XRL) method for trading financial stocks. In this paper, we proposed to employ SHapley Additive exPlanation (SHAP) on a popular deep reinforcement learning architecture viz., deep Q network (DQN) to explain an agent's action at a given instance in financial stock trading. To demonstrate the effectiveness of our method, we tested it on two popular datasets namely, SENSEX and DJIA and reported the results.

## 1. Introduction

Artificial intelligence (AI) has permeated many areas of our daily lives over the last few decades. Unfortunately, as these models become more powerful and adaptable, they also become more opaque, rigid, and black boxes. In machine learning, interpretability refers to the ability of a model to describe the entire behavior of the model, from input to output. The more formal term for this is Explainable Artificial Intelligence (XAI), which applies to all forms of AI. First, we need to consider clear psychological factors. If people don't believe in a model or prediction, they don't use it. Transparency is emphasized as a key factor in increasing system users' and user acceptance. This is essential for using the model or system [1].

---

[*] Corresponding Author

The goal of XAI research in the context of machine learning and deep learning is to look inside this "black box" and understand more about the reasoning behind the decisions and actions made by the algorithm. XAI addresses bias in machine learning and debugging, in addition to providing tools to assist with responsibility and trust. Machine learning algorithms are frequently subject to bias or human error since they are ultimately developed with human input (i.e., humans in the loop) [2, 3]. The explanations offered by XAI-enabled algorithms may identify possible flaws or problems with this architecture.

XAI methods can be categorized based on two factors [1].
i) Intrinsic or post-hoc (based on when the information is extracted): An intrinsic model is an ML model built to be interpretable or self-explanatory at the time of training by limiting the complexity of the model, e.g. decision trees. In post-hoc analysis, the model after training is recreated by creating a second, simpler model, to provide explanations for the original model. This is carried out for surrogate models. Intrinsic models usually offer accurate explanations, but due to their simplicity, their prediction performance suffers. As they need to keep the accuracy of the original model intact, it is harder to derive satisfying and simple explanations. ii) Global or Local (based on the scope of the problem being addressed): Global techniques explain the overall behavior of the model, while local techniques explain a particular decision. Global models attempt to explain all the logic of a model by examining the structure of the model. Local explanations try to address the following kinds of questions: Why does the model make a certain prediction or decision for a case or for a group of cases? Global interpretation techniques lead users to trust a model, while local techniques lead to confidence in a prediction.

Reinforcement learning is a branch of machine learning and decision-making science developed by Sutton and Barto [4] that focuses on understanding how to act in situations to maximize rewards. You need to maximize the numerical reward signal to relate situations to behaviors and learn to behave optimally. Environmental models (the world of agents where agents live and interact), with policies, rewards, value functions, and options, are the four basic sub-elements of a reinforcement learning system.

Recent advances in Deep Learning (DL) for learning feature representations have had a significant impact on RL, and the fusion of the two techniques (known as Deep RL) has produced amazing results in many different domains. RL is typically used to solve optimization problems when the system has a lot of states and a complicated stochastic structure. Training agents to play Atari games based on raw pixels [5, 6], complex real-world robotics problems like manipulation [7], or recommendation systems [8], are notable examples.

Although Deep Reinforcement Learning (DRL) has emerged as a promising machine learning approach for sequential decision-making problems, it is still in its infancy for high-stakes tasks like autonomous driving, finance, defense, or medical applications. In such cases, a learned policy, for example, must be interpretable so that it can be inspected before deployment [9]. Researchers proposed and tested various RL interpretation methods in a variety of applications. While the overall objective of RL interpretation is to make RL policies more understandable, each method has its own unique set of goals, problems, limitations, and challenges [10].

The method used when trading stocks is important when making a financial investment. However, in a turbulent market, it is difficult to create an ideal stock trading environment. Deep reinforcement learning has been used to solve this problem. DRL is used to train in the live market and learn advanced trading techniques to maximize returns on investment. Regarding how RL agents trade, there are still unanswered questions. The task of stock trading has been extensively researched using DRL. The block-box architecture of deep neural networks makes it difficult to understand a trading strategy based on DRLs. The risk of entrusting a neural network with important trading funds is high because the network is a black box [11].

In this paper, we propose an explanation methodology of the actions taken by an agent in a trained DQN model in the context of financial stock market trading using SHapley Additive exPlanation (SHAP) [12]), which is been applied to the indices (SENSEX, DJIA)

The rest of the paper is organized as follows: Section 2 discusses the literature survey work. Section 3 describes the methodologies which are essential to understanding the XRL for stock trading, namely, the Deep Q-Network (DQN) and SHAP. In section 4, the dataset description is provided. Section 5 presents the experimental setup along with the results obtained for each of the datasets. Finally, section 6 concludes the paper.

## 2. Literature Survey

Previous research on explainable RL covers the literature that attempts to explain MDP-based agents. These older works usually provide local explanations because the explanation is for a question about an agent's action.

Elizalde et al. [13] developed explanations using the concept of "relevant variables" in a factored state of an MDP. Explanations were incorporated and presented when a trainee made a poor decision, with the main goal of explaining things to human system learners. Each action has a relevant variable that is selected by an expert; in this case, the explanation is the relevant variable. Later, Elizalde et al. [14], Elizalde and Sucar [15] expanded on this work to automatically construct explanations depending on the usefulness of a state variable to the policy selecting the action.

Work carried out by Khan et al. [16] on minimally satisfactory explanations for MDPs was influenced by the relevant variable explanations. Here, the long-term effects of an optimal action are considered when building the explanation. Three generic templates served as the foundation for the explanations.

Wang et al. [17] examined how the explanations offered by MDP-based agents alter the concept of "trust". Trust in human-robot teams was assessed in tests that were influenced by partially observable MDP-based explanations. Because the rise in trust was measured by self-report, it is unclear whether it was brought on by a genuine understanding of the system.

The literature review that has been conducted by various researchers in the field of XRL is summarized in Table 1. It gives us an understanding of the environment and algorithms that have been presented in the past, and the column of explanation type indicates how an agent's action is explained.

Previous studies have looked at RL with reward decomposition, but they put more emphasis on accelerating learning than on providing explanations. Reinforcement learning (RL) agents' decisions can be explained via reward decomposition. In this method, rewards are decomposed into the sum of reward types that have semantic significance so that actions can be compared in terms of trade-offs between the types [18].

## 3. Methodology

Recent deep reinforcement learning applications in the financial markets take into account discrete or continuous state and action spaces and use one of three learning methodologies: actor-only, actor-critic, or critic-only [28].

### 3.1 Deep Q- Network

Deep-Q Networks (DQN) were first introduced for Atari 2600 video games [5]. A DQN uses a neural network to roughly represent a state-value function in a Q-Learning framework.A memory table Q[s,a] is constructed in Q-learning to store Q-values for each conceivable combination of states (s) and action (a) . The Q-Value function, which the agent learns, provides the anticipated total return for a given state and action pair. As a result, the agent must behave in a way that maximizes this Q-Value function [29].

For the stock market, input multiple frames and output the status value for each action [29]. This is typically used in combination with experience replays to store episode stages in memory for out-of-policy learning using randomly selected samples from replay memory. Q networks are often optimized for frozen target networks that are updated every k steps with the latest weights (k is a hyperparameter). The latter improves training stability by reducing the temporary vibrations caused by target changes. By dealing with the autocorrelation that results from online learning, the first approach is closer to the problem of supervised learning.

**Table 1.** Summary of literature review

| Reference | Task/ Environment | Decision process | Algorithms | Explanation type | Target |
|---|---|---|---|---|---|
| [19] | Planning +strategy games | POMDP | IMPALA | Images (local) | Experts |
| [20] | Robotics(grasping), and navigation | MDP | PPO | Diagram (state plot & image slider) | Experts |
| [18] | Game(grid and landing) | MDP | HRA, SARSA, Q-learning | Diagrams (Local) | Experts, Users, Executives |
| [21] | Games (OpenAI) | Both | PG,DQN, DDPG, A2C, SARSA | Diagrams, Text (Local) | Experts, Users, Executives |
| [22] | Multiagent Navigation, Prey and predator | POMDP | DDPG | Diagrams (Local) | Experts |
| [23] | Robotics(grasping) | MDP | DDPG, HER, HRL | Diagrams (Global) | Experts, Developers |
| [24] | Game | MDP | DQN | Diagrams | Experts |
| [25] | Arcade games | POMDP | Q-Learning | images(Local) | Users |
| [26] | Arcade games | POMDP | Q-Learning | images(Local) | Users |
| [27] | Robotics | MDP | Soft Q-learning | Diagrams(local) | Experts |

## 3.2 SHapley Additive exPlanation

The game theoretic optimal Shapley values are the foundation of the SHapley Additive exPlanation (SHAP) approach [12], which is used to explain each instance's prediction. The explanation of an instance is given as a linear function of features, and it is a member of the class of models known as additive feature attribution methods. An instance of feature serves as players in a coalition in the SHAP explanation approach, which is based on coalitional game theory. Shapley values are utilized for fair distribution of prediction which is analogous to payouts in coalitional game theory.

Let f be the original prediction model that needs to be explained, and let g be the explanation model. Here, we concentrate on local approaches that aim to explain a prediction f(x) based on a single input x. Explanation models frequently use simplified inputs x′ that map to the original inputs using a mapping function $x = h_x(x')$. When z′≈x′, local approaches attempt to ensure that $g(z') \approx f(h_x(x'))$. Because $h_x$ is particular to the current input x, it should be noted that $h_x(x') = x$ even though x′ may have less information than x. The explanation offered by SHAP has the following mathematical representation

$$g(z'_j) = \phi_0 + \sum_{i=1}^{M} \phi_j z'_j$$

where M is the coalition vector representing maximum size of the coalition, j is the contribution of features to the final output in the presence of feature j, and g is the explanation model $g(z_j')$ which is a local surrogate model of the original model f(x), and which aids in the interpretation of the original model.

In order to calculate a feature Shapley value, all feasible feature value combinations are weighted and added together.

$$\phi_j(N) = \sum_{S \subset N \setminus \{i\}} \frac{|S|!\,(|N| - |S| - 1)!}{|N|!} (v(S \cup \{i\}) - v(S)),$$

,w here S is the subset of features used in the model, N is the set of all possible permutations of features, x is the vector of feature values for the instance that needs to be explained, v(S) is the prediction for comparatively small feature values in set S compared to features not in set S.

The most popular learning strategy, known as critic-only learning, trains an agent on a single stock or asset to solve a discrete action space problem employing, for example, Deep Q-learning (DQN) and its modifications [30-32].

The critic-only technique uses a Q-value function to learn the optimal action-selection policy that maximizes the predicted future reward given the current state.

We build an environment that is similar to the real trading market prior to training a DRL agent so that it can participate in interaction and learning. Practical trading requires consideration of a number of factors, such as past stock prices, present shareholdings, technical indicators, etc. Our trading agent needs to obtain this information from the environment in order to make the necessary decisions.

To maximize returns, stock traders aim to trade stocks over time. A discrete period of time known as an episode is used to simulate this process using MDP. In a single episode, an agent consecutively interacts with an environment. Every episode is broken into time steps, which are denoted by $t \in \{0,1,2,\cdots,T\}$, and each episode represents one trading day. When time steps traveling sequentially reach t=T, a single trading episode is considered to be complete. In particular, this paper limits its analysis of trading frequency to daily.

The stock trading environment is obtained from gym-anytrading [33]. The action space describes the actions that an agent must take to interact with its environment. Action taken at time t denoted by $a_t \in A$, where A is the action space and $a_t \in \{0,1\}$, where 0 represents selling of stocks and 1 represents buying of stocks. The reward function $r(s_t, a_t, s_{t+1})$ a strategy that encourages the agent to discover more efficient strategies. When an action is taken, the stock's overall value changes, and this is referred to as the reward.

Figure 1 depicts a schematic illustration of the proposed methodology, with input data consisting of all stocks from 2 stock market indexes (SENSEX and DJIA). In the preprocessing step following things are carried out: removed rows for the corresponding missing values, changed the data type of date feature from string to date type and set it as an index, date is sorted in ascending order from oldest to latest. The data set is divided into training data, which is considered from January 1, 2014, to May 31, 2021 (1836 occurrences), and test data, which is considered from June 01, 2021, to June 21, 2022 (253 instances). Train data is used to train DQN. Predictions are made using test data, and the given model predicts the maximum cumulative rewards depending on the action chosen by an agent. Reward predictions are made using a sliding window with actions from the past 30 days. The opening, low, high, and closing prices of each stock are among the inputs that are sent to DQN for both training and trading (or testing). SHAP provides an explanation of the prediction of the maximum cumulative rewards. The user is shown these explanations via waterfall plots.

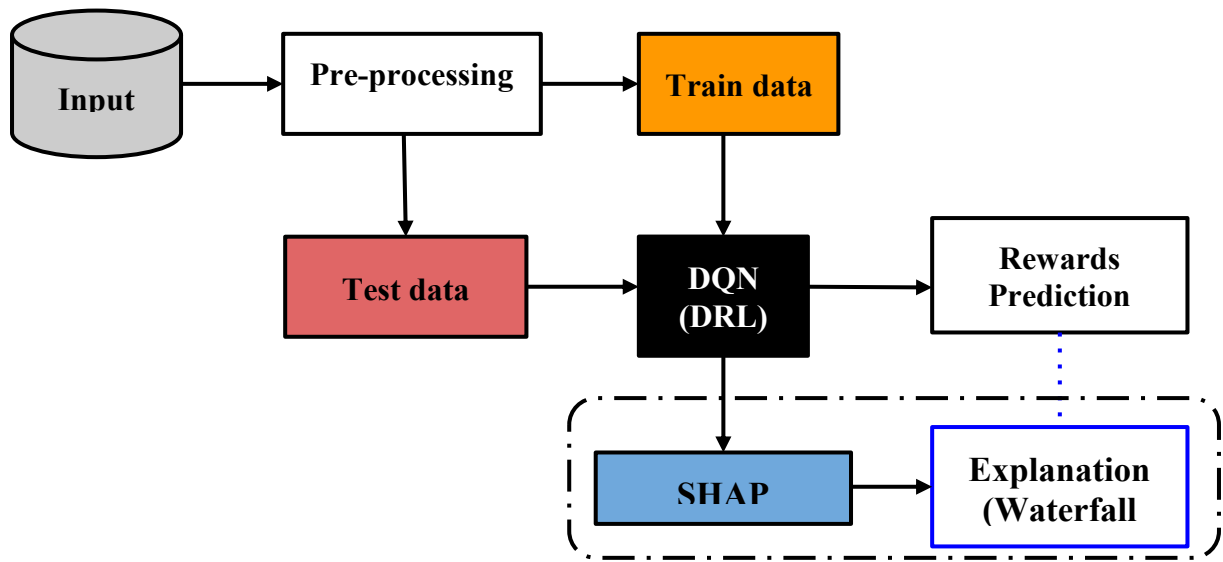

**Fig 1.** Pipeline of XRL for financial stock trading.

## 4. Description of Datasets

In this section, stocks from 2 stock market indices (SENSEX and DJIA) are presented and are downloaded using the Yahoo Finance API [34]. Each stock market index has a total of 30 companies trading. The features that are common to all stocks are as follows: opening price, closing price, low price, high price achieved by companies, and the number of shares traded during the day (Volume). The acronym for these combined features is written as OHLCV.

### 4.1 SENSEX

SENSEX stands for Stock Exchange Sensitive Index, and it is a stock market index for the Bombay Stock Exchange (BSE). It is one of the oldest market indices in India and is operated by Standard & Poor's (S&P). It is also known as BSE 30. The SENSEX is a market index composed of 30 financially stable, successful, and well-established companies [35].

Following are the 5 criteria that companies need to fulfill to be part of SENSEX
- It must be listed on the BSE.
- High Liquidity.
- Mid-sized or large companies.
- Generate substantial revenue from core business activities.
- Keep a balanced sector weight of companies.

The names of companies which are part of SENSEX are as follows:[36]: Asian Paints Limited, Axis Bank, Bajaj, Auto Limited, Bajaj Finserv Limited, Bajaj Finance Limited, Bharti Airtel Limited, HCL Technologies Limited, Housing Development Finance Corporation Limited, HDFC Bank Limited, Hindustan Unilever Limited, ICICI Bank Limited, IndusInd Bank Limited, Infosys Limited, ITC Limited, Kotak Mahindra Bank Limited, Larsen & Toubro Limited, Mahindra & Mahindra Limited, Maruti Suzuki India Limited, Nestle India Limited, NTPC Limited Oil & Natural Gas Corporation Limited, Power Grid Corporation of India Limited, Reliance Industries Limited, State Bank of India, Sun Pharmaceutical Industries Limited, Tata Steel Limited, Tata Consultancy Services Limited, Tech Mahindra Limited, Titan Company Limited, UltraTech Cement Limited.

## 4.2 Dow Jones Industrial Average

The Dow Jones Industrial Average (DJIA), commonly referred to as "Dow Jones" or simply "Dow", is one of the most popular and well-known stock market indexes. It measures the daily stock market movements of 30 US-listed companies listed on NASDAQ or the New York Stock Exchange (NYSE) [37]. The Dow Jones is named after Charles Dow, who created the index in 1896 with his trading partner Edward Jones.

The names of companies which are part of DJIA are as follows [38]. 3M, American Express, Amgen, Apple, Boeing, Caterpillar, Chevron, Cisco Systems, Coca-Cola, Disney, Dow, Goldman Sachs, Home Depot, Honeywell, IBM, Intel, Johnson & Johnson, JP Morgan Chase, McDonald's, Merck, Microsoft, Nike, Procter & Gamble, Salesforce, Travelers, UnitedHealth, Visa, Walgreens, and Walmart are just a few of the companies that have made the list.

## 5. Results and Discussion

As part of the model input, previous time steps are added. For example, the current time-step will be included along with the three previous time-steps in a sliding window of size 4. We can easily train an accurate value function with a sliding window of 30 inputs.

The explanations obtained by SHAP for the rewards predicted by DQN are presented in Table 2. The actions taken by an agent over the last 30 days, with a sliding window of 30, are being used to calculate the reward of a test instance of the present day based on the action performed. The test window advances to predict the next state's reward based on the last 30 day's action taken by an agent. Mathematically, explanation for an instance $(x_t)$ on the current day's reward prediction can be represented as follow

$$\xi(x_t) = (r(s_{t-29}, a_{t-29}), r(s_{t-28}, a_{t-28}), \cdots, r(s_t, a_t)),$$

Where $\xi$ is the explanation provided by SHAP which is used for explaining prediction of reward value obtained for current day t and s denotes state. For each state there are 2 possible actions of either buying stocks or selling stocks. The base rate is the average of all reward predictions made by the model and is represented by E (f(x)), where f(x) is the reward predictions made by the model for a given instance. A single waterfall plot is provided for each day of the test instance as an explanation of the actions performed by an agent. In the waterfall plot which is shown by the Features t as the output and this is representing each state $(s_t)$ where $t \in \{0,1,2,\cdots,29\}$. The (30 -i)$^{th}$ day of test data represented by the waterfall plot's i$^{th}$ feature. For e.g.: Feature 0 denotes day 30. Plots with a red or blue bar, respectively, display buy action or sell actions for a specific date. Each bar represents SHAP values for each of the days considered in sliding window. Top 10 contributing days (or features) are chosen, and because of their minimal contribution to prediction, 21 features are combined into 1 feature.

Throughout the paper following conventions are followed:
- If f(x) <E (f(x)), then either action performed by an agent will result in loss for the current day.
- If f(x)>E (f(x)), then either of the action performed by an agent will result in profit for the current day.
- Action performed till i$^{th}$ day is shown by OHLC plot and explanation of i$^{th}$ instance is explained by Waterfall plots.

**Table 2.** Summary of explanation for stock trading.

| Environment | Stock Trading |
|---|---|
| Agent (or Algorithm) | DQN |
| Type of Explanation | Local [Explanation obtained by Waterfall plot ] |
| What is to be explained? | Reward that an agent expects to receive from a buy or sell action on a particular day based on the actions taken in the past 30 |
| Process of explanation | The window advances by one date, and the most recent action done by an agent is considered. Each date yields a single explanatory plot. |
| Decision Process | MDP |
| Target | Experts, Executives |

The list of all hyperparameters used in DQN training is shown in Table 3. For training data, two hidden layers of 50 neurons each are used. Two output neurons are used because it is based on a discrete action space with two values. Each episode of DQN's training consists of 100 training epochs over the course of 30 episodes. A discount factor (γ) of 0.95 is chosen and used to update the target (or reward) value.

**Table 3.** Hyperparameters used for training DQN

| Attributes | Value |
| --- | --- |
| Discount factor ($\gamma$) | 0.95 |
| Exploration rate ($\varepsilon$) | 1.0 |
| Min Exploration rate | 0.01 |
| Learning rate | 0.005 |
| Batch size | 32 |
| Number of hidden neurons | 50 |
| Epochs | 100 |
| Episode | 30 |
| Activation function | ReLU |
| Neurons in output layer | 2 |
| Loss function | Mean Squared Error (MSE) |
| Optimizer | Adam |

## 5.1 SENSEX

The prediction plot shown in Fig. A.1 includes OHCV values of the test data and also the agent's action, till the tenth instance of SBIN. Here the 40$^{th}$ test instance is explained and the algorithm in this case decides to buy stocks ($a_{10}$=1), as the 30 days date is included in test data. Fig. A.2. illustrates an explanation for January 25, 2022 $\xi(x_{10})$, taking into account features from December 26, 2021 to January 24, 2022 $(s_{29}, s_{28}, \cdots, s_0)$, for which the model predicts a reward of 10.743. Action performed for $s_{26}$ is a sell action, with expected reward of 1.94 greatly influences the current day's reward prediction. Even though a positive reward for the buy action, this action has resulted in loss for the trader, hence the trader should not buy action on January 25, 2022.

The prediction plot shown in Fig. A.3 includes the OHCV values of the test data and also the agent's action, till the 20$^{th}$ instance of Reliance stock. Here the 50$^{th}$ test instance ($x_{50}$) is explained and the algorithm in this case decides to buy stocks ($a_{20}$=1), as the first 30 test data are included in the sliding window of size 30. Fig. A.4. illustrates an explanation for August 11, 2021 $\bigl(\xi(x_{50})\bigr)$, taking into account features from June 30, 2021 to August 10, 2021 $(s_{29}, s_{28}, \cdots, s_0)$, for which the model predicts a reward of 773.826. Action performed for $s_{15}$ is a buy action, with an expected reward of 5.6 greatly influences the current day's reward prediction. This positive reward for the buy action will resulted in profit for the trader, hence the trader should buy stocks of Reliance on August 11, 2021.

The prediction plot shown in Fig. A.5 includes the OHCV values of the test data and also the agent's action, till the 20[th] instance of BAJFINANCE stock. Here the 50th test instance $(x_{50})$ is explained and the algorithm in this case decides to buy stocks ($a_{20}$=1), as the first 30 test data are included in the sliding window of size 30. Fig. A.6. illustrates an explanation for August 11, 2021$(\xi\ (x_{50}))$, taking into account features from June 30, 2021 to August 10, 2021$(s_{29}, s_{28}, \cdots, s_0)$, for which the model predicts a reward of -2821.986. Action performed on $s_6$ is a sell action, with an expected reward of 62.6. Buy action performed on $s_4, s_2, and\ s_5$, which has expected positive rewards of 38.53, 19.48 and 19 respectively seems to dominate over $s_6$, thus greatly influences the current day's reward prediction. Even though a negative reward for the buy action is obtained, this action has resulted in profit for the trader, hence the trader should consider buying stocks of BAJFINANCE on August 11, 2021.

The prediction plot shown in Fig. A.7 includes the OHCV values of the test data and also the agent's action, till the 21[th] instance of BAJAJFINSV stock. Here the 50th test instance $(x_{51})$ is explained and the algorithm in this case decides to sell stocks ($a_{21}$=0), as the first 30 test data are included in the sliding window of size 30. Fig. A.8. illustrates an explanation for August 12, 2021$(\xi\ (x_{51}))$, taking into account features from June 30, 2021 to August 11, 2021$(s_{29}, s_{28}, \cdots, s_0)$, for which the model predicts a reward of -1302.222. Action performed on $s_{26}$ is a buy action, with an expected reward of -155.62. Sell action performed on $s_2, and\ s_{24}$, which has expected positive rewards of 128.67 and 126.53 respectively seems to dominate over $s_{26}$, thus greatly influences the current day's reward prediction. Even though a negative reward for the sell action is obtained, this action has resulted in profit for the trader, hence the trader should consider selling stocks of BAJAJFINSV on August 12, 2021.

The prediction plot shown in Fig. A.9 includes the OHCV values of the test data and also the agent's action, till the 30[th] instance of NESTLEIND stock. Here the 60th test instance $(x_{60})$ is explained and the algorithm in this case decides to sell stocks ($a_{30}$=0), as the first 30 test data are included in the sliding window of size 30. Fig. A.10. illustrates an explanation for September 09, 2021$(\xi\ (x_{60}))$, taking into account features from July 28, 2021 to September 08, 2021$(s_{29}, s_{28}, \cdots, s_0)$, for which the model predicts a reward of 7633.834. Action performed on $s_{12}$ is a sell action, with an expected reward of 211.03. Even though a positive reward for the sell action is obtained, this action has resulted in loss for the trader, hence the trader should not consider selling stocks of NESTLEIND on September 09, 2021.

## 5.2 DJIA

The prediction plot shown in Fig.B.1 includes OHCV values of the test data and also the agent's action, till the 19th instance of AAPL. Here the 49th test instance is explained and the algorithm in this case decides to buy stocks ($a_{39}$=1), as the 30 days date is included in test data. Fig. B.2. illustrates an explanation for August 10, 2021 $\xi(x_{49})$, taking into account features from June 30, 2021 to August 09, 2021 ($s_{29}, s_{28}, \cdots, s_0$), for which the model predicts a reward of 121.308. Action performed for $s_{28}$ is a sell action, with expected reward of -1.35 does not seem to influence the current day's reward prediction. Buy action performed on $s_5, s_{26}, s_{11}$ and $s_{12}$ which has expected positive rewards of 0.55, 0.26, 0.25 and 0.23 respectively seems to dominate over $s_{28}$, thus greatly influences the current day's reward prediction. Even though a positive reward for the buy action, this action has resulted in loss for the trader, hence the trader should not buy action on August 10, 2021.

The prediction plot shown in Fig. B.3 includes the OHCV values of the test data and also the agent's action, till the 21th instance of AXP stock. Here the 51th test instance ($x_{51}$) is explained and the algorithm in this case decides to sell stock ($a_{21}$=0), as the first 30 test data are included in the sliding window of size 30. Fig. B.4. illustrates an explanation for August 13, 2021 $(\xi(x_{51}))$, taking into account features from July 01, 2021 to August 12, 2021 ($s_{29}, s_{28}, \cdots, s_0$), for which the model predicts a reward of -18.572. Action performed for $s_{30}$ is a buy action, with an expected reward of 0.59. Sell action performed on $s_5, s_{26}, s_{11}$ and $s_{12}$ which has expected positive rewards of 0.55, 0.26, 0.25 and 0.23 respectively seems to dominate over $s_{28}$, thus greatly influences the current day's reward prediction This negative reward for the sell action will result in profit for the trader, hence the trader should sell stocks of AXP on August 13, 2021.

The prediction plot shown in Fig. B.5 includes the OHCV values of the test data and also the agent's action, till the 15th instance of BA stock. Here the 45th test instance ($x_{45}$) is explained and the algorithm in this case decides to buy stock ($a_{20}$=1), as the first 30 test data are included in the sliding window of size 30. Fig. B.6. illustrates an explanation for August 05, 2021 $(\xi(x_{45}))$, taking into account features from June 23, 2021 to August 4, 2021 ($s_{29}, s_{28}, \cdots, s_0$), for which the model predicts a reward of -14.343. Action performed on $s_{22}$ is a buyaction, with an expected reward of 1.92, which greatly influences the current day's reward prediction. Even though a negative reward for the buy action is obtained, this action has resulted in loss for the trader, hence the trader should not consider buying stocks of BA on August 05, 2021.

The prediction plot shown in Fig. B.7 includes the OHCV values of the test data and also the agent's action, till the 34th instance of CAT stock. Here the 64th test instance ($x_{64}$) is explained and the algorithm in this case decides to sell stocks ($a_{64}$=0), as the first 30 test data are included in the sliding window of size 30. Fig. B.8. illustrates an explanation for September 1, 2021$(\xi(x_{64}))$, taking into account features from July 22, 2021 to August 31, 2021$(s_{29}, s_{28}, \cdots, s_0)$, for which the model predicts a reward of -46.542. Action performed on $s_{16}$ is a sell action, with an expected reward of -0.77. Sell action performed on $s_5, s_6$ and $s_{28}$, which has expected positive rewards of 0.75, 0.62 and 0.43 respectively does not seem to have much of an influence in reward prediction. Even though a negative reward for the sell action is obtained, this action has resulted in loss for the trader, hence the trader should not consider selling stocks of CAT on September 1, 2021.

The prediction plot shown in Fig. B.9 includes the OHCV values of the test data and also the agent's action, till the 10[th] instance of CSCO stock. Here the 40th test instance $(x_{40})$ is explained and the algorithm in this case decides to sell stocks ($a_{40}$=0), as the first 30 test data are included in the sliding window of size 30. Fig. B.10. illustrates an explanation for July 29, 2021$(\xi(x_{40}))$, taking into account features from June 04, 2021 to July 28, 2021$(s_{29}, s_{28}, \cdots, s_0)$, for which the model predicts a reward of -17.459. Action performed on $s_6$ is a buy action, with an expected reward of 0.2 is the most important feature, but action performed on $s_{26}$ with an expected reward of -0.09 seems to overshadow contribution of $s_6$ which influences the current day's reward prediction. Even though a negative reward for the sell action is obtained, this action has resulted in profit for the trader, hence the trader should consider selling stocks of CSCO on July 29, 2021.

## 6. Conclusion and Future Scope

In this paper, we proposed how explainability is applied using reinforcement learning in the stock market using SHAP for stocks of Sensex and DJIA individually for every stock which is first of its kind. We provided explanations for actions taken by the DQN agent using plots given in this section.

SENSEX and DJIA, two major stock market indices, are evaluated for explainability using SHAP based on agent-provided reward predictions. In order to improve the explanation, in future more technical indicators will be added as features along with three actions (-1=sell, 0=hold, and 1=buy). To explain agent behavior for stock market indices, it may also be extended to continuous action space using DRL models such as Deep Deterministic Policy Gradient (DDPG) [39], Advantage Actor-Critic (A2C) [40], and others.

*Acknowledgments*

*The authors are thankful to Mr. Rajiv Ramachandran for explaining us the stock trading ideas and guiding us during his tenure as Senior Domain Expert at IDRBT.*

# Appendix A: Prediction and Explanation plots of SENSEX

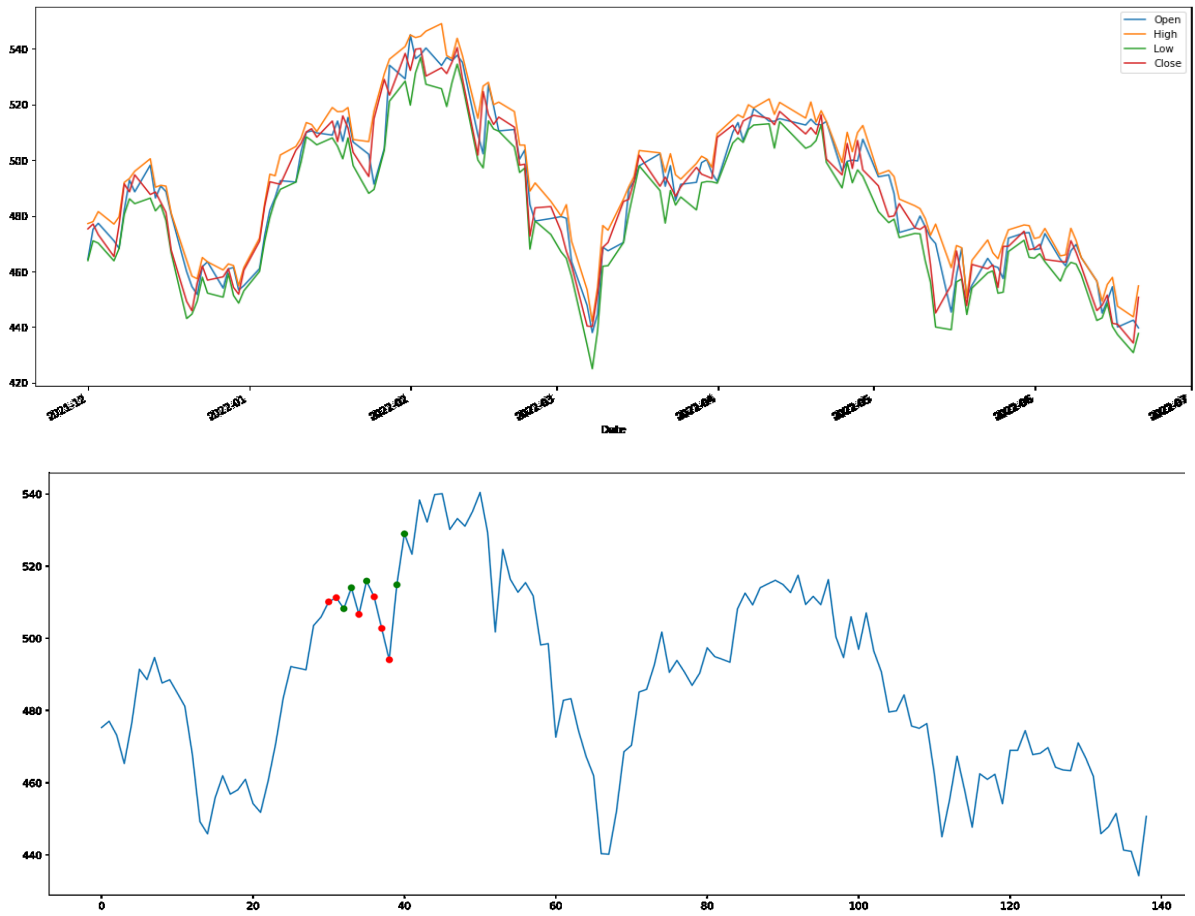

Fig. A.1. Predictions made by DQN for the 10th instance on SBIN stock.

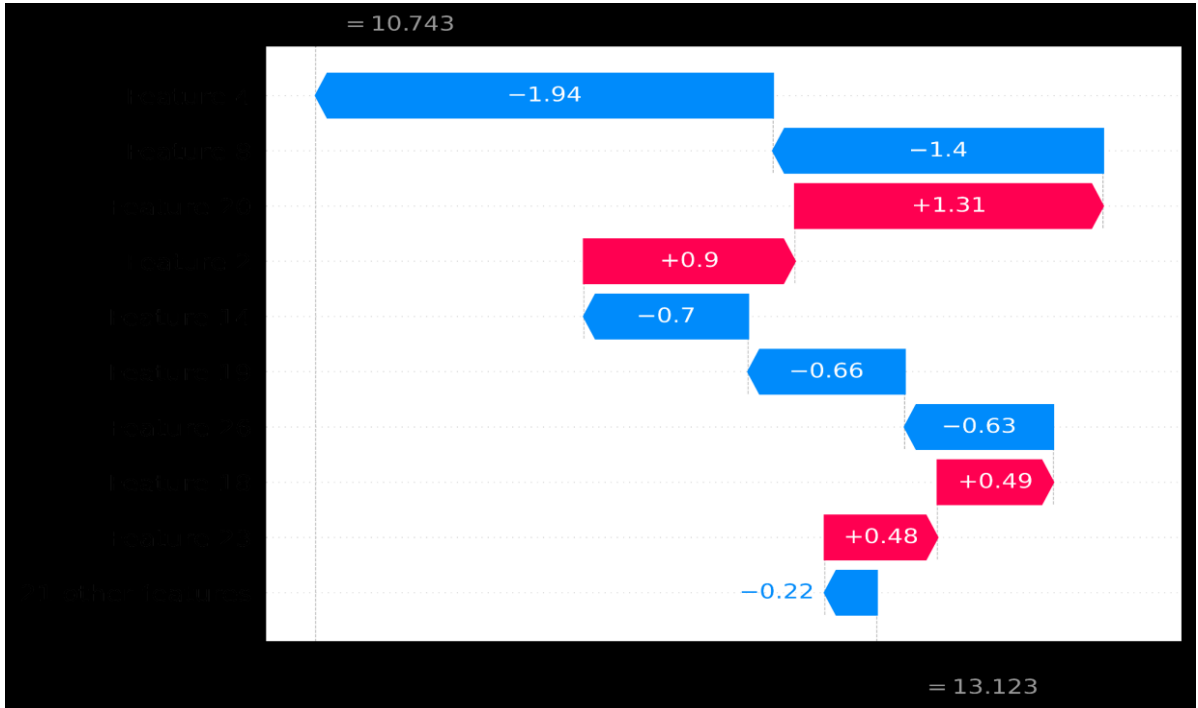

Fig. A.2. Explanation of reward predictions of SBIN stock for buy action.

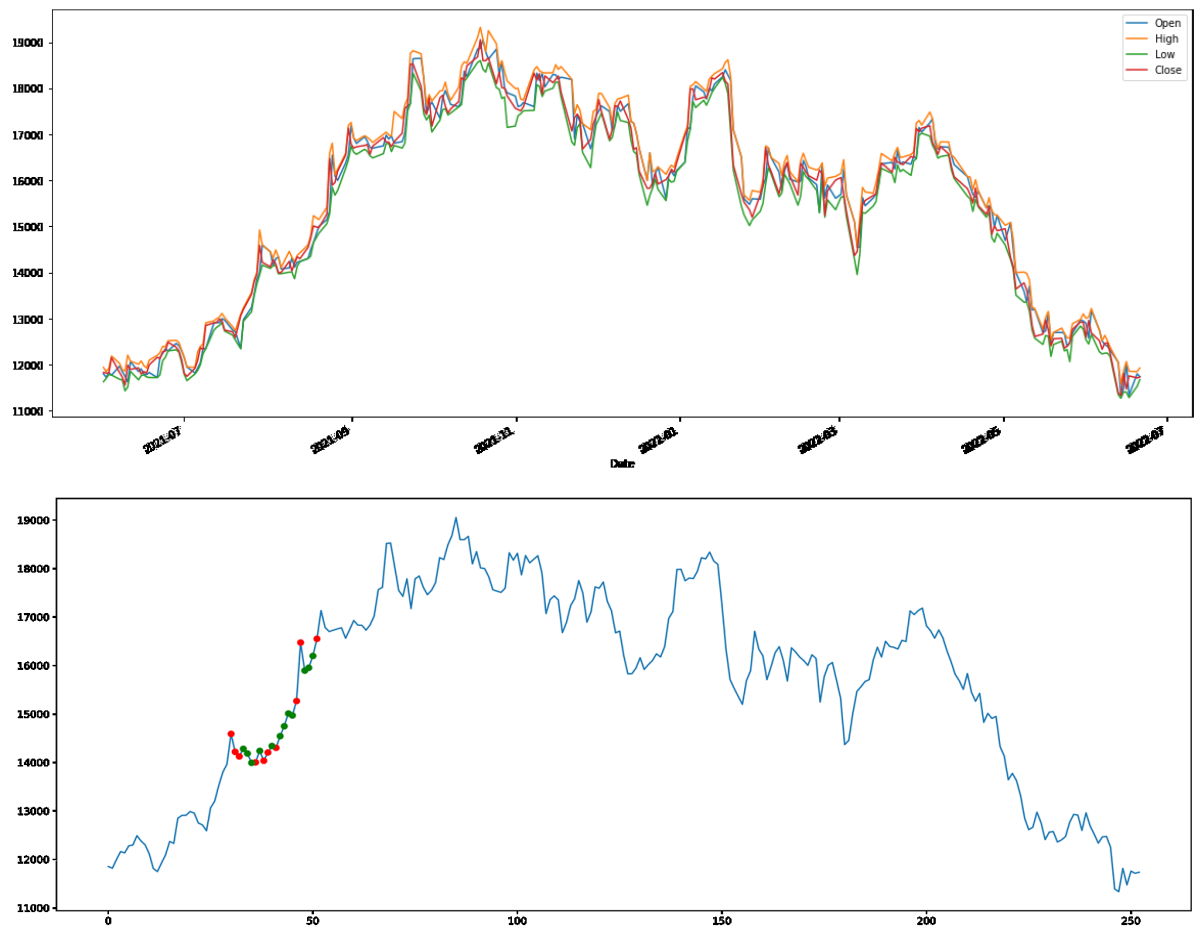

Fig. A.3. Predictions made by DQN for the 20[th] instance of Reliance stock.

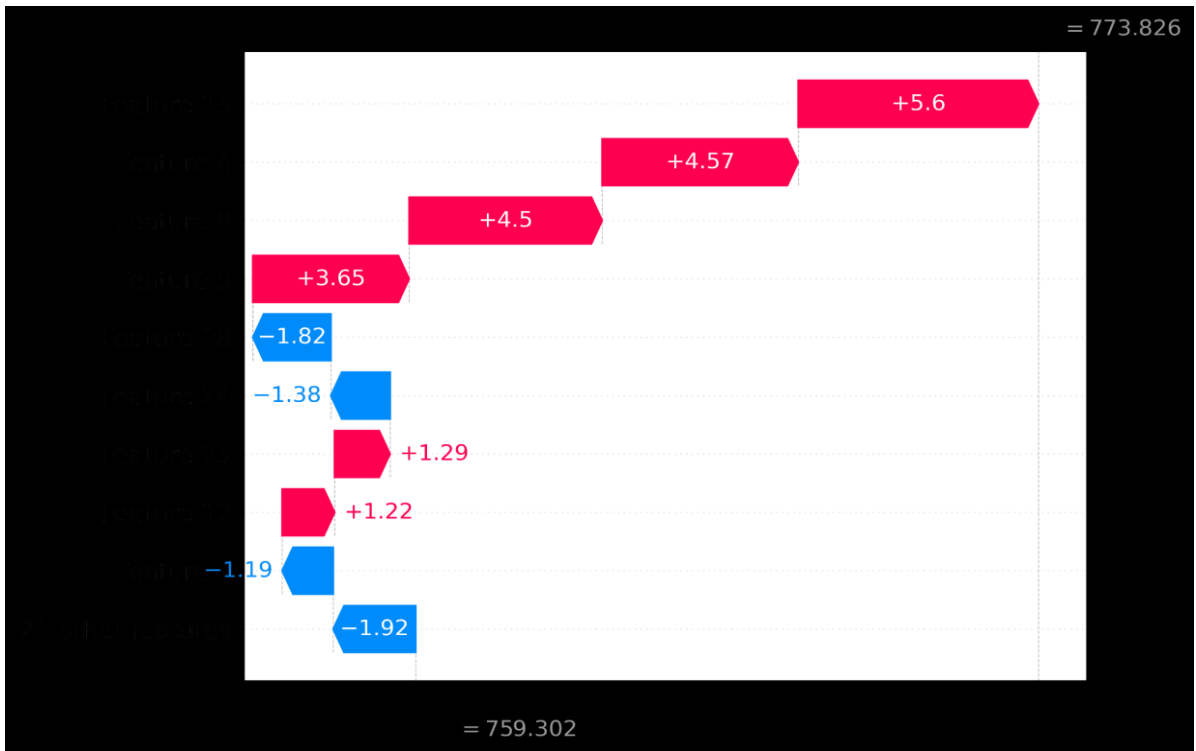

Fig. A.4. Explanation of reward predictions of Reliance stock for buy action.

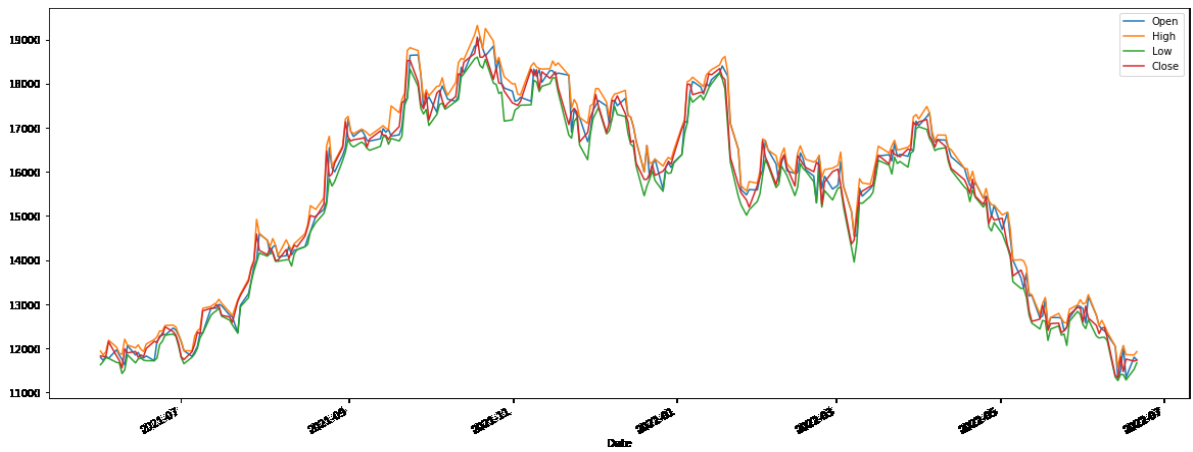

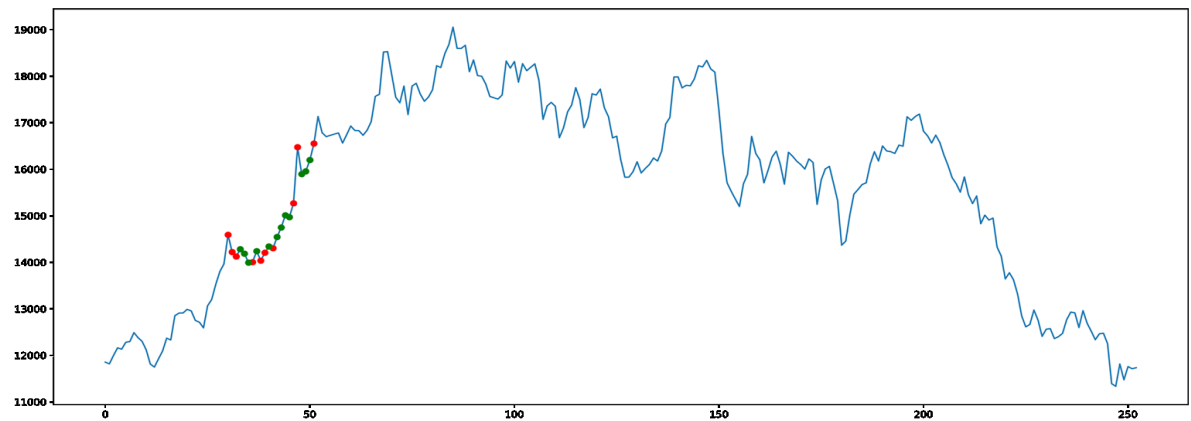

Fig. A.5. Predictions made by DQN for the 20[th] instance of BAJFINANCE stocks.

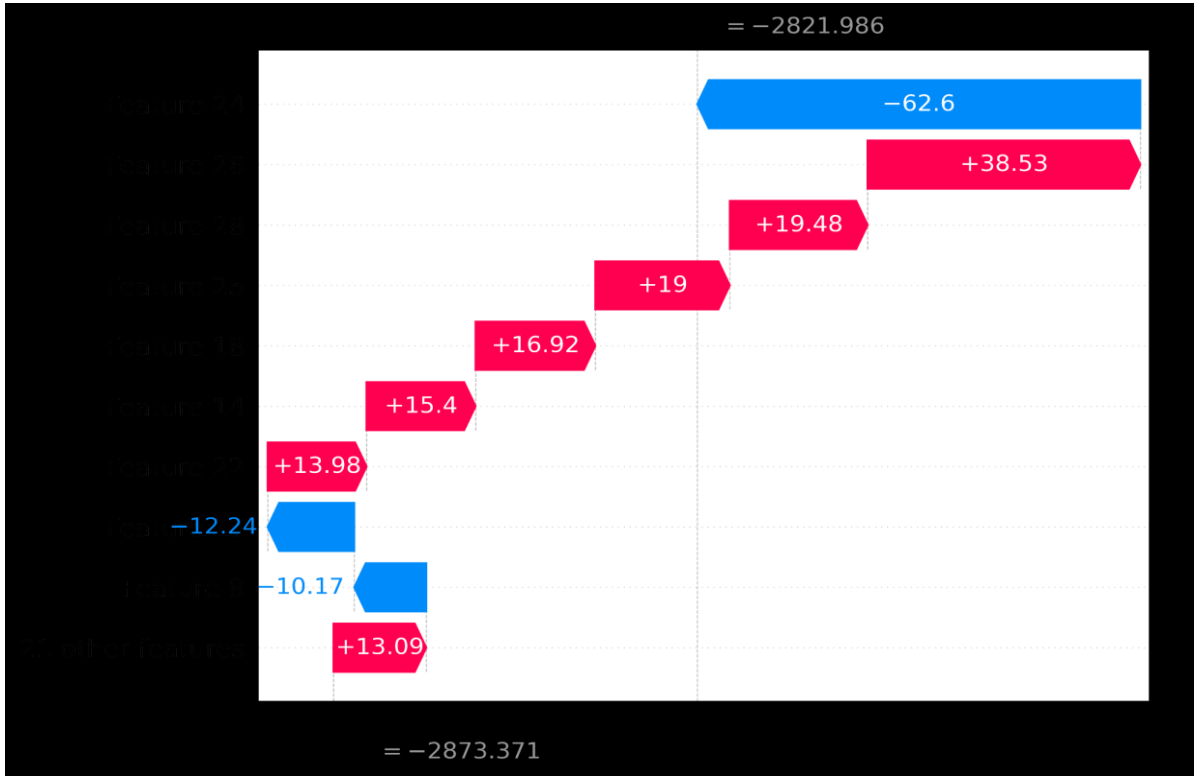

Fig. A.6. Explanation of reward predictions of BAJFINANCE stock for buy action.

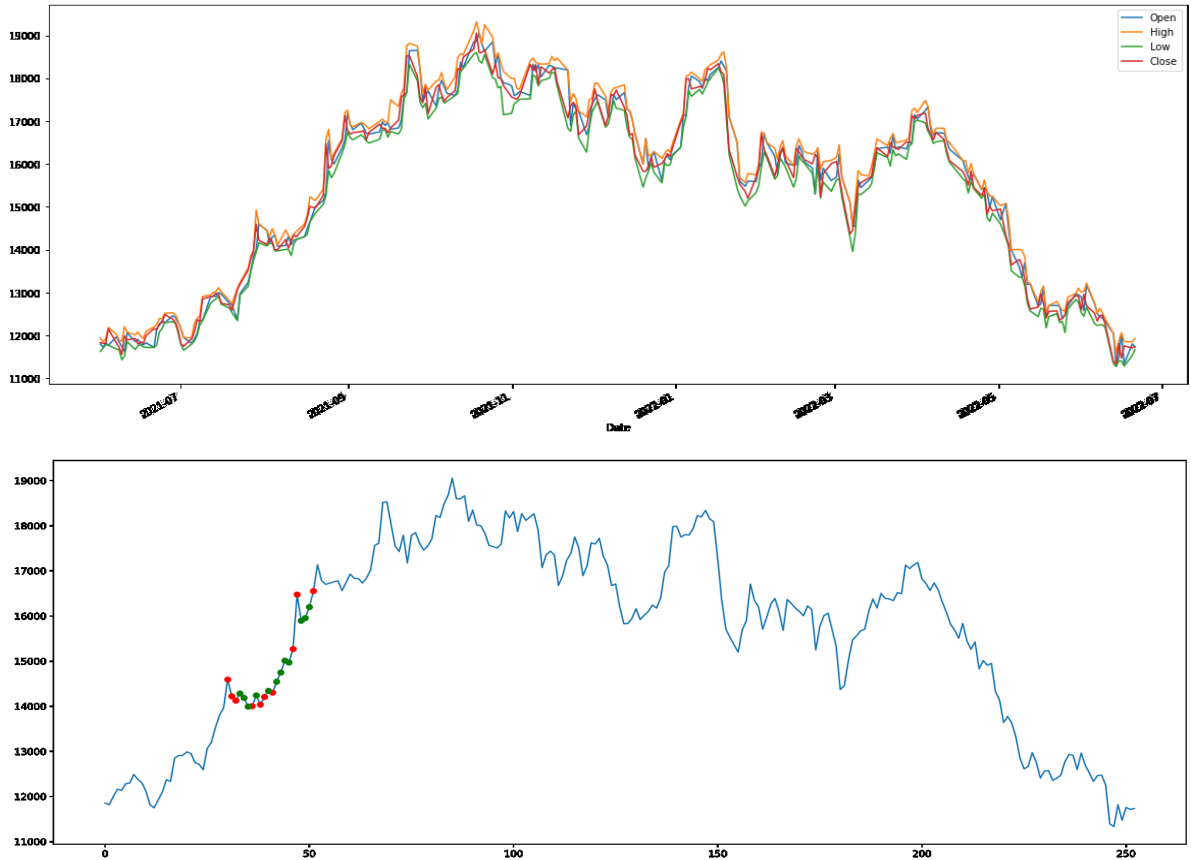

Fig. A.7. Predictions made by DQN for the 21$^{st}$ instance of BAJAJFINSV stocks.

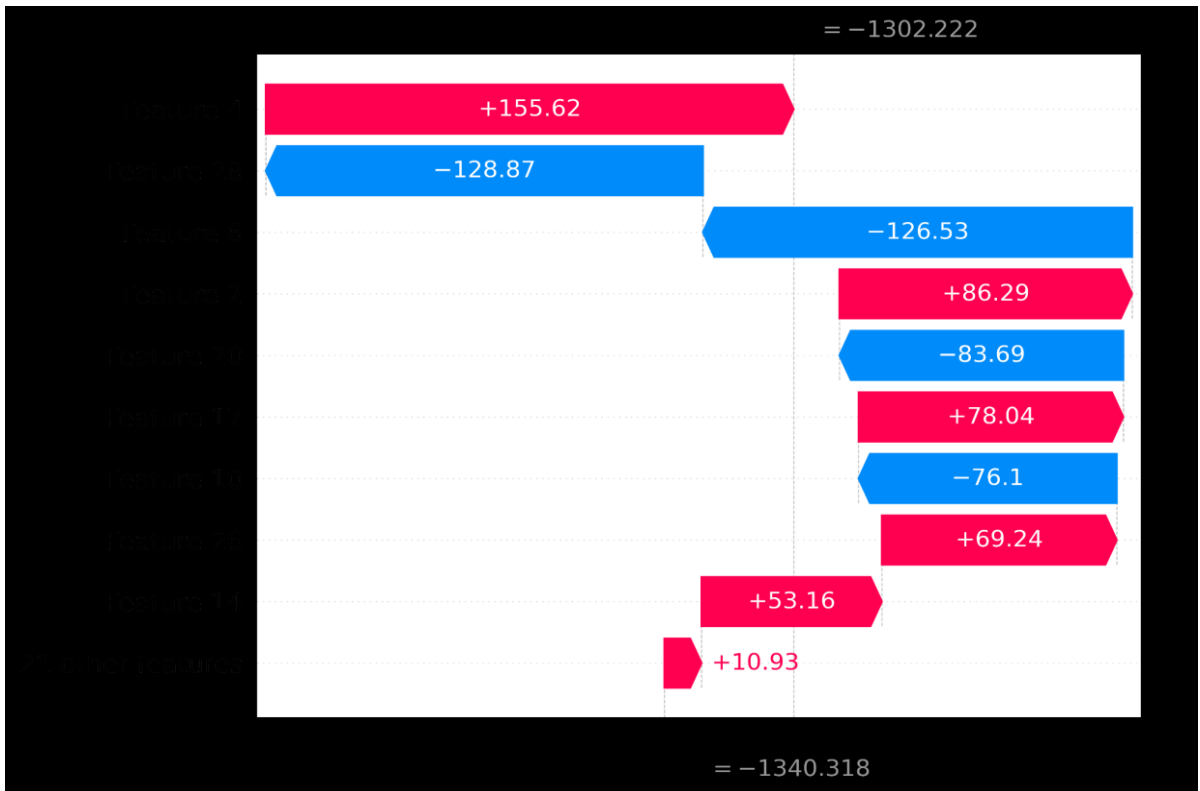

Fig. A.8. Explanation of reward predictions of BAJAJFINSV stock for sell action.

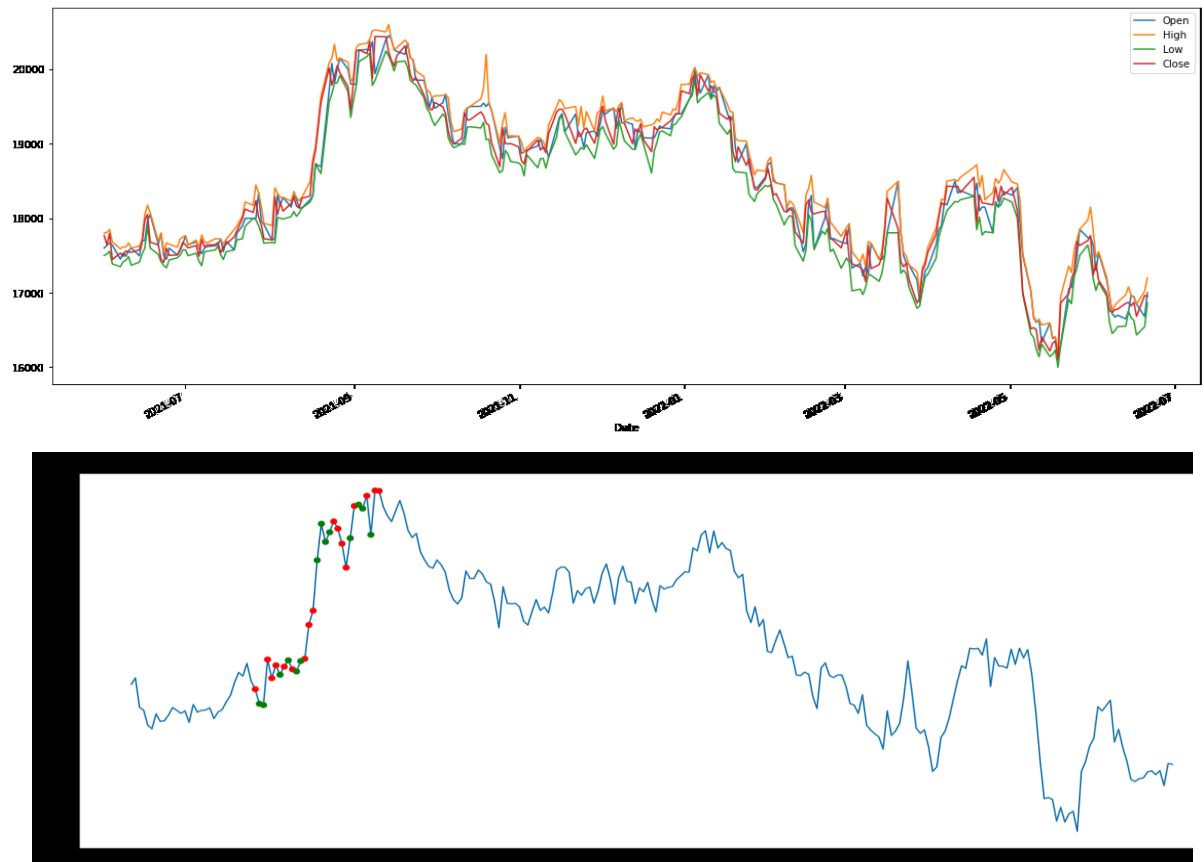

Fig. A.9. Predictions made by DQN for 30[th] instance of NESTLEIND stocks.

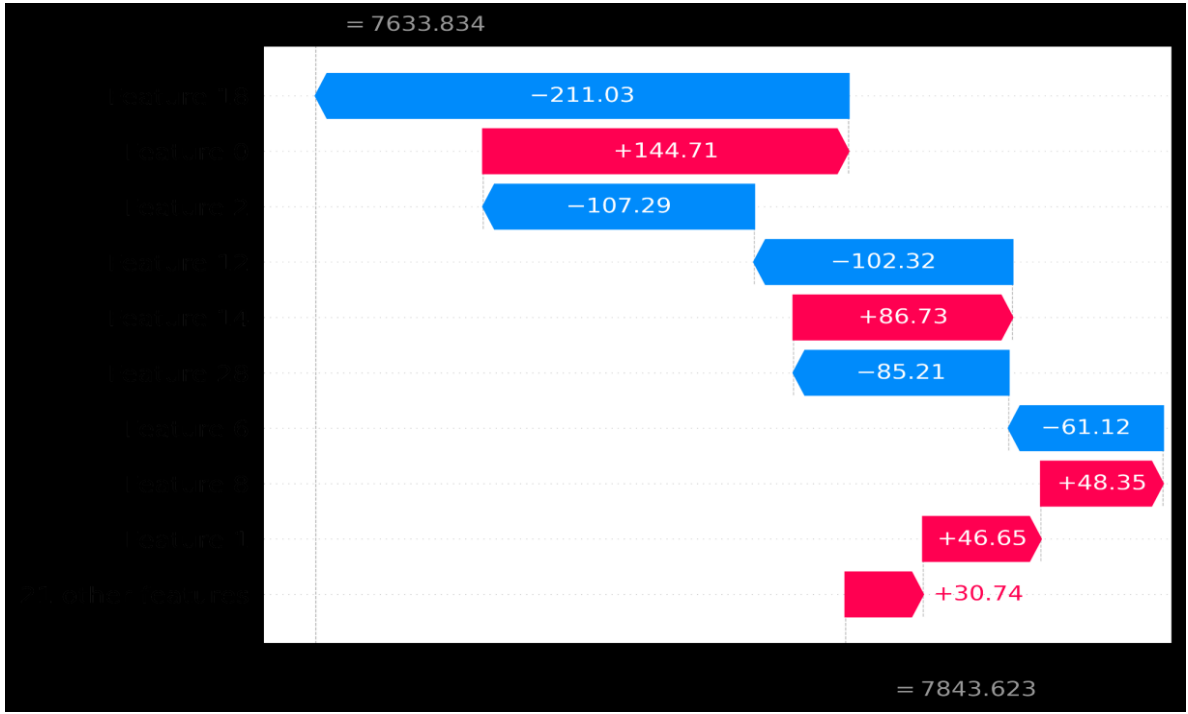

Fig. A10. Explanation of reward predictions of NESTLEIND stock for sell action.

## Appendix B: Prediction and Explanation plots of DJIA

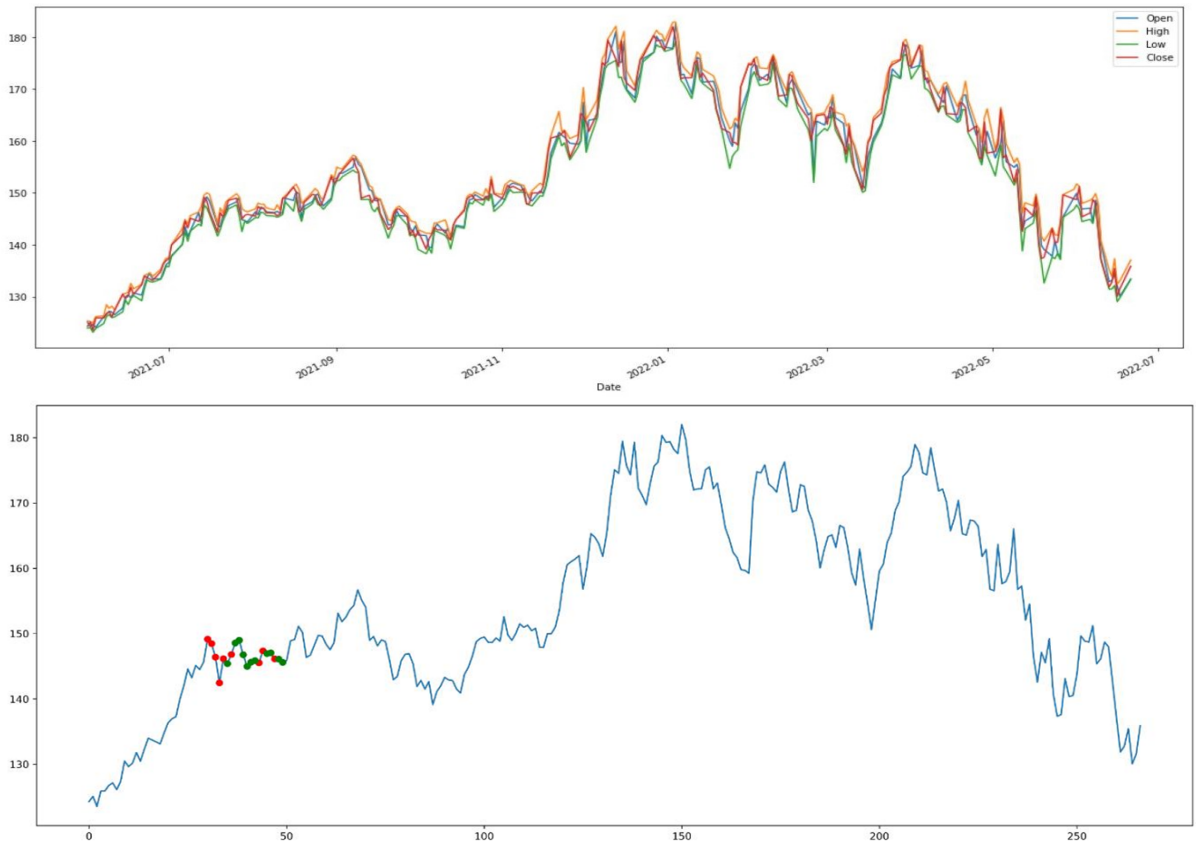

Fig. B.1. Predictions made by DQN for the 19th instance of AAPL stock.

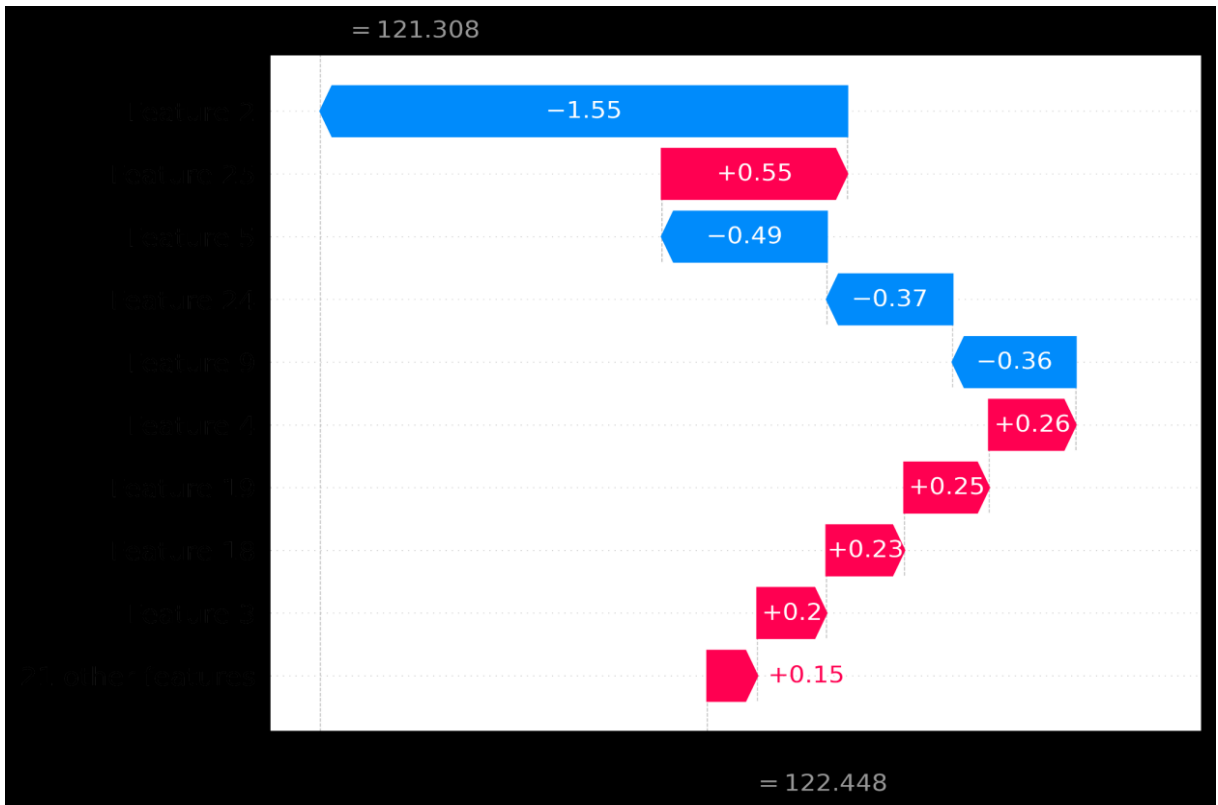

Fig. B.2. Explanation of reward predictions of AAPL stock for buy action.

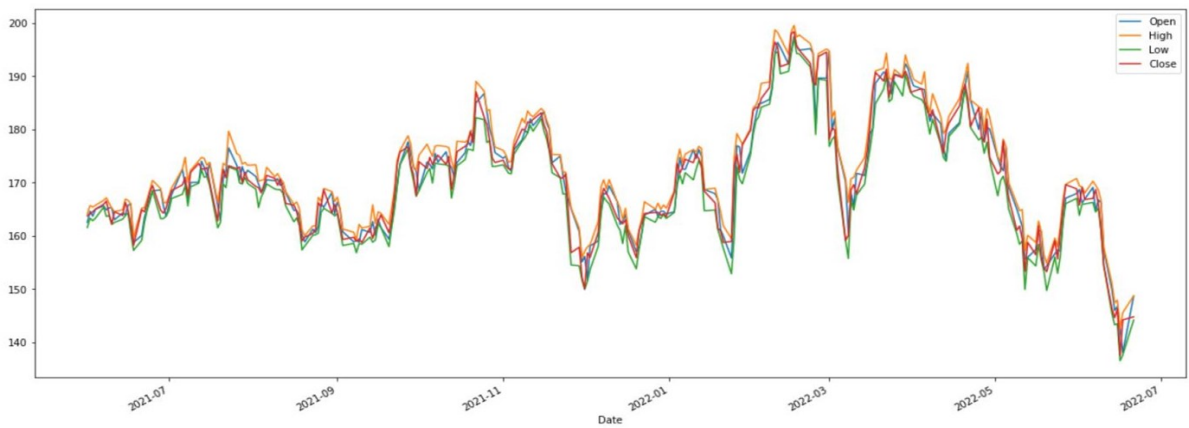

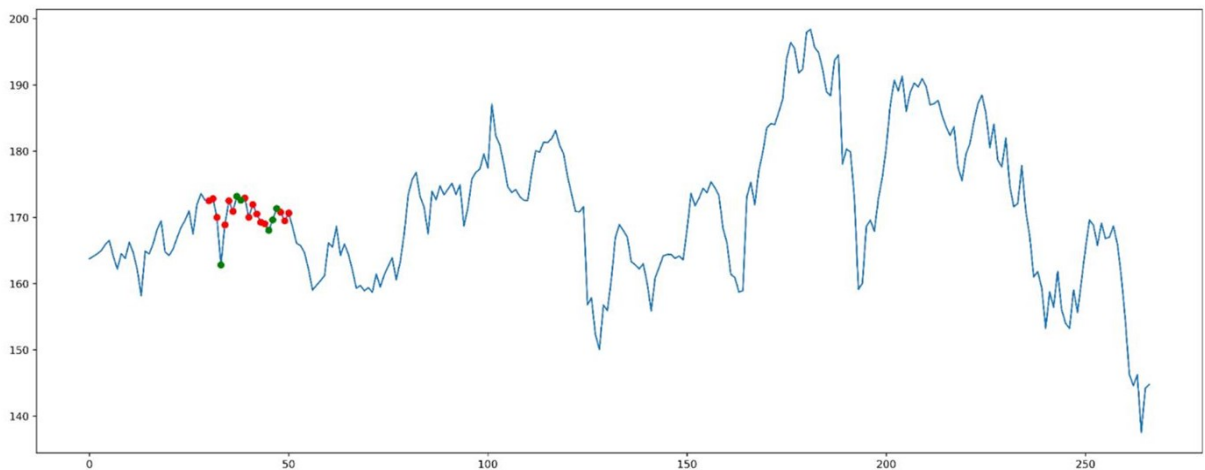

Fig. B.3. Predictions made by DQN for the 21st instance of AXP stock.

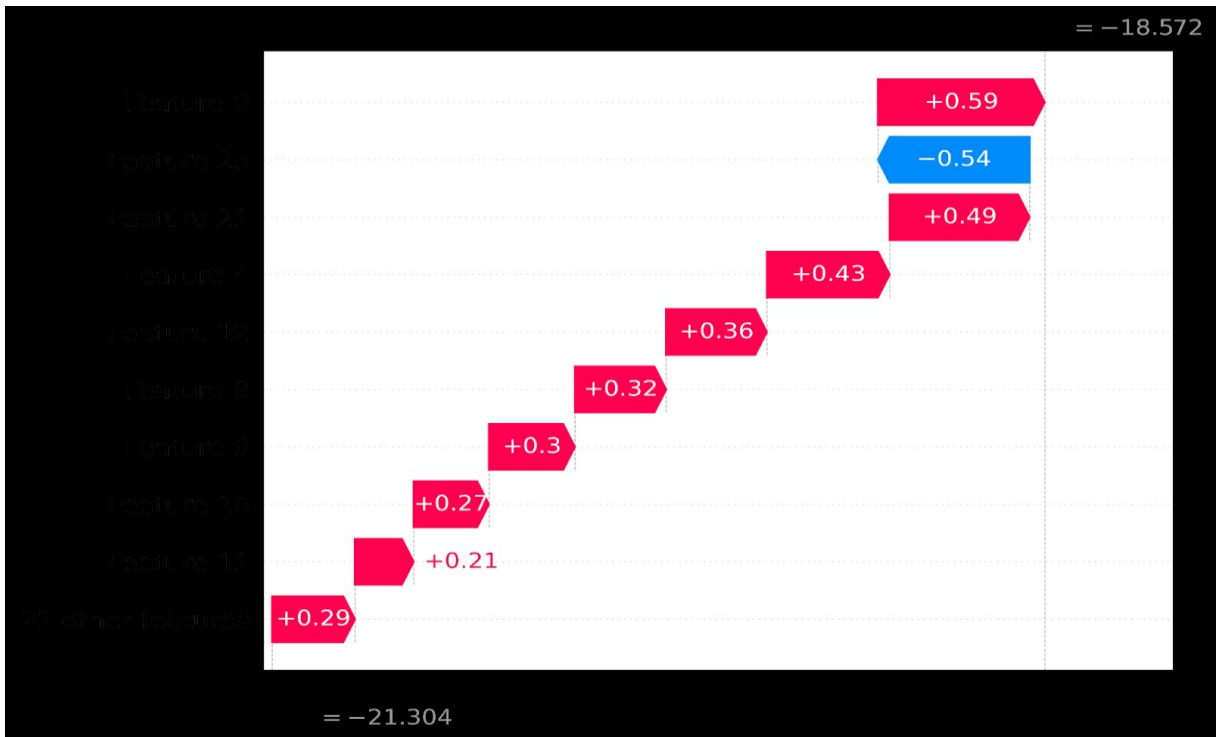

Fig. B.4. Explanation of reward predictions of AXP stock for sell action.

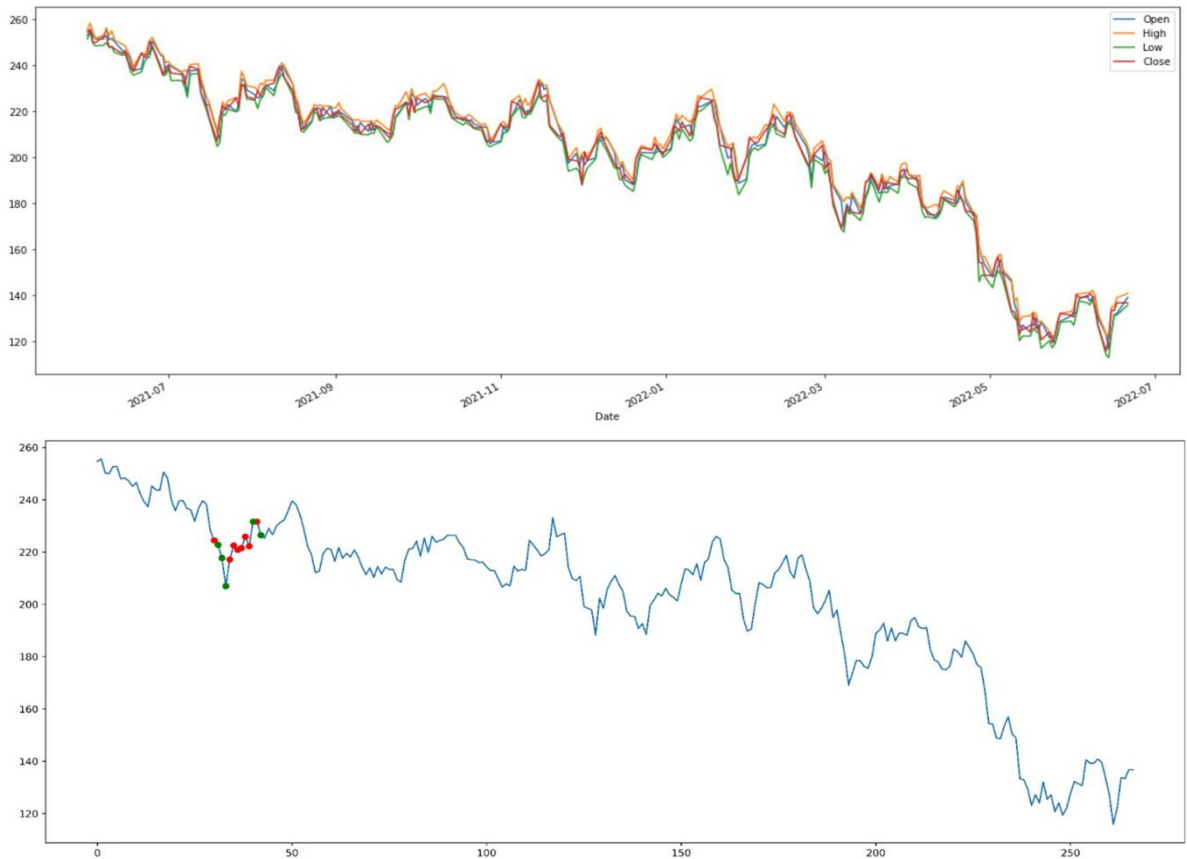

Fig. B.5. Predictions made by DQN for the 15th instance of BA stock.

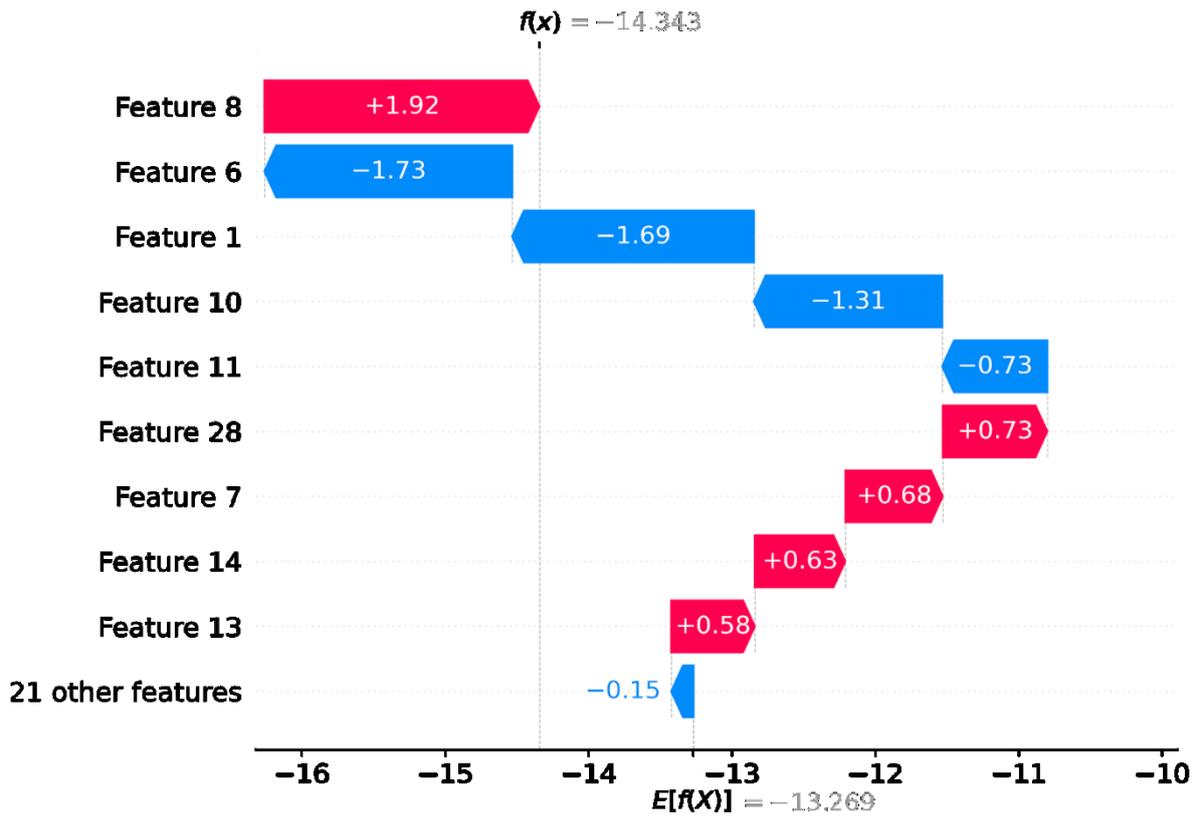

Fig. B.6. Explanation of reward predictions of BA stock for buy action.

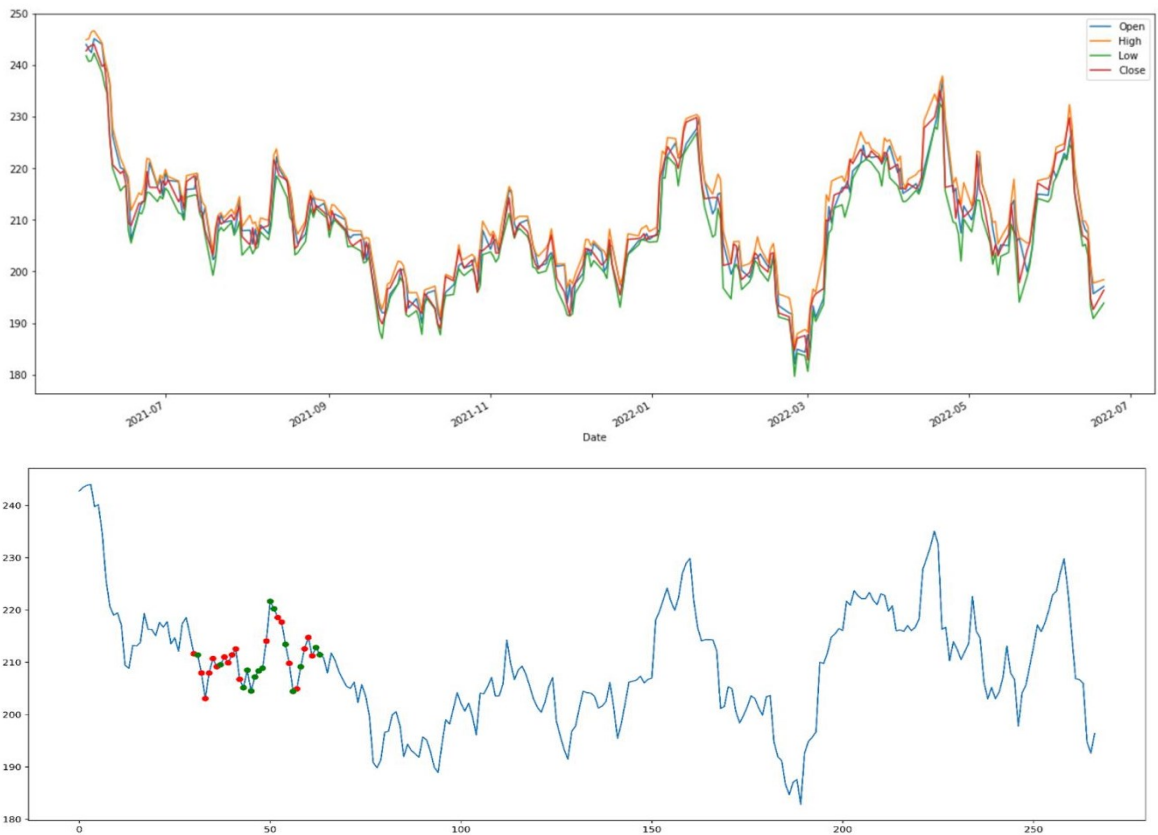

Fig. B.7. Predictions made by DQN for the 34[th] instance of CAT stock.

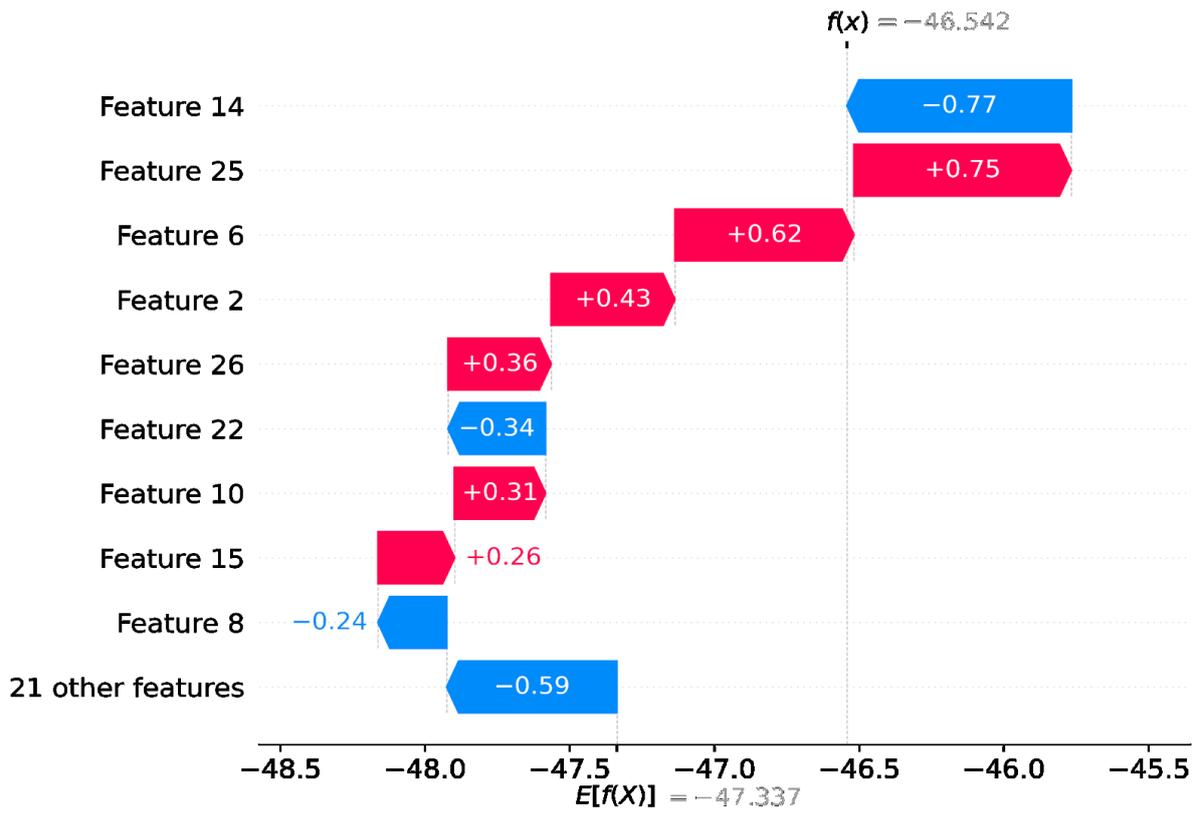

Fig.B.8. Explanation of reward predictions of CAT stock for sell action.

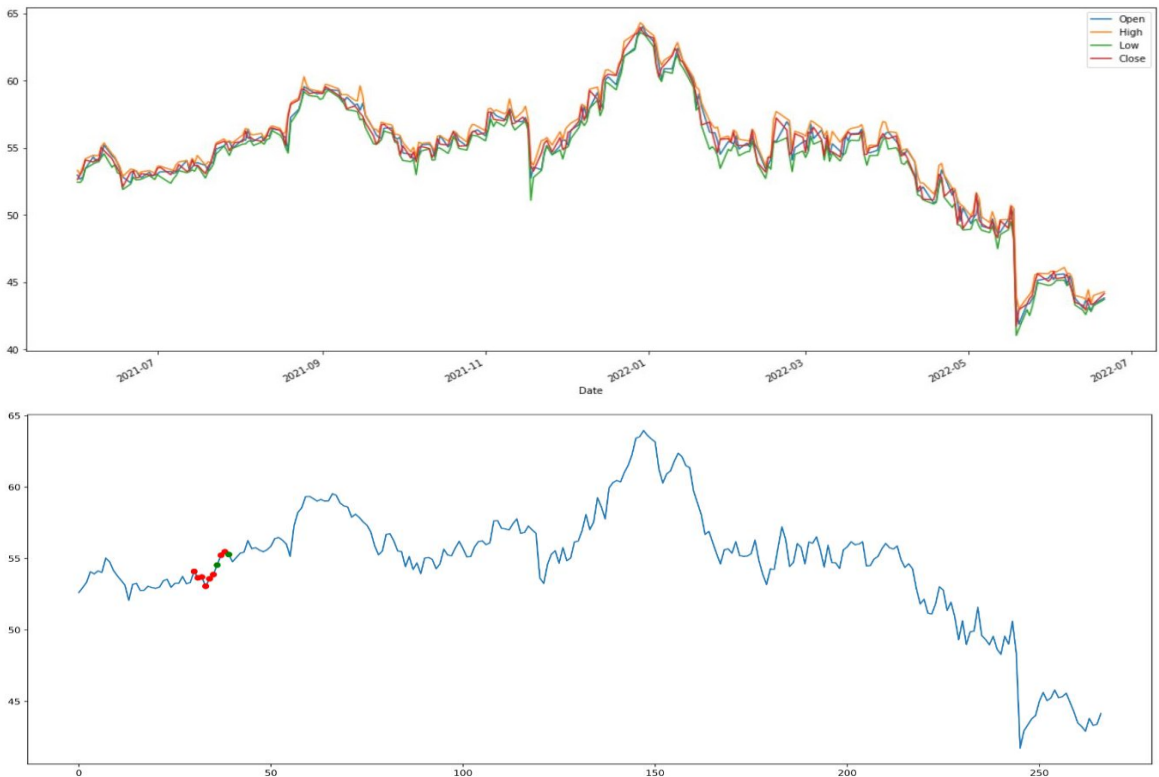

Fig. B.9. Predictions made by DQN for the 10th instance of CSCO stock.

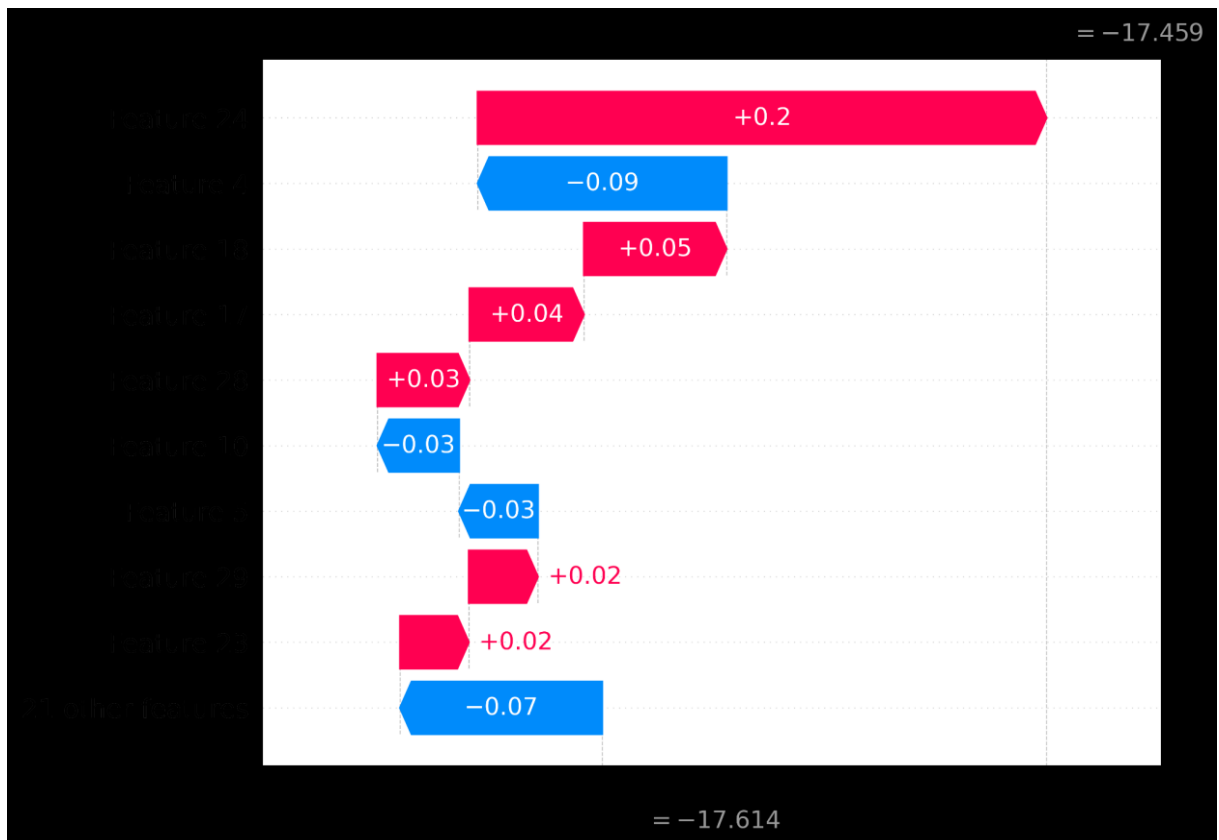

Fig.B.10. Explanation of reward predictions of CSCO stock for sell action.